\documentclass{article}
\usepackage[natbibapa]{apacite} 

\usepackage{arxiv}

\usepackage[utf8]{inputenc} 
\usepackage[T1]{fontenc}    
\usepackage{hyperref}       
\usepackage{url}            
\usepackage{booktabs}       
\usepackage{amsfonts}       
\usepackage{nicefrac}       
\usepackage{microtype}      
\usepackage{tabularx}
\usepackage{fancyvrb}       
\usepackage{graphicx}
\usepackage{doi}
\usepackage{authblk} 

\title{Evaluation of pre-trained models for pedagogical assessment of novel AI-assisted educational questions}
\author[1]{
Michael Lawrence Castanares\thanks{\texttt{michael@predictivesystems.ai}}
}
\author[1]{Princess Ventures}
\author[1,2]{Allan Tan}
\affil[1]{Predictive Systems Inc, Ortigas, Pasig City, Philippines, 1605}
\affil[2]{Better Labs Oy, Helsinki, Finland}

\date{}

\renewcommand{\shorttitle}{Evaluating Models for Pedagogical Assessment of AI Educational Questions}

\hypersetup{
    colorlinks=false,   
    pdfborder={0 0 0}    
}

\begin{document}
\maketitle

\begin{abstract}
The surge in AI-assisted generation of educational materials has outpaced our capacity to validate their pedagogical quality. Automated evaluation using Bloom Classifier models is a promising approach to assess educational materials at scale. These models show high accuracy within-distribution dataset (IID Dataset). However, applying the same models to new out-of-distribution (OOD) datasets such as AI-assisted generated questions could show performance degradation. To identify robust classifiers under dataset shift, we evaluated traditional Machine Learning (ML), transformer, and Large Language models on the Bloom level classification task. We also explored feature-engineering strategies incorporating  NLP metrics, appending the learning objectives as part of the input, and text splicing to stabilize OOD performance. Our baseline tests show that TFPOS-IDF ML models perform poorly on OOD (Macro F1-score 0.48) compared to BERT (0.55) and LLMs (0.79). Text splicing improved macro F1-score performance of ML and BERT models (0.59 and 0.62, respectively). Appending the learning objectives with the input increased model performance on specific dataset. Model retraining provided the largest improvement across models and datasets. Overall, these findings highlight the trade-off on the use of pre-trained models with novel AI-assisted educational and how strategic feature enhancements help address loss in performance.
\end{abstract}

\keywords{Revised Bloom Taxonomy, Machine Learning, Artificial Intelligence}

\section{Introduction}

The surge in AI-assisted educational questions(AEQ) has outpaced our capacity to validate their pedagogical quality \citep{xing2024, scaria2024, wang2025}. While LLMs are able to produce learning content at scale, ensuring their alignment with educational frameworks such as the Revised Bloom Taxonomy and Common European Framework of Reference (CEFR) is a concern. Traditional expert-validation of questions is costly, time-consuming, and prone to human biases \citep{alkhuzaey2024, wang2025}. 

Quality validation covers the alignment of materials with the Revised Bloom's Taxonomy for cognitive depth and CEFR for language proficiency. Educational materials should capture both learning objectives and student's level of knowledge against a defined difficulty. Moreover, the materials should allow progression from low-order thinking tasks (e.g., remember, understand, and apply) to higher-order cognitive tasks (e.g., analyze, evaluate, and create).

Studies in Natural Language Processing (NLP) developed models capable of detecting Bloom signals (i.e., Bloom trigger Verbs) \citep{mohammed2020, waheed2021} and readability levels of text at scale \citep{arase2022}. These models show high accuracy on IID. Whether these models deliver the same accuracy with novel AEQ remains unknown.

AI-generated questions are verbose and may lack explicit linguistic signals (e.g., Bloom Verbs) found in human-curated training datasets used in previous models \citep{mohammed2020, yahya2012, lau2025}. Feature-based models heavily rely on detecting these Bloom trigger verbs for classification.  We hypothesize that text verbosity in AI-generated questions dilutes the Bloom signals resulting to under-performance of the models.

To test this hypothesis, we examined the performance of different pre-trained models (SVM, XGBoost, BERT transformers, and Large Language Models) with AEQ and across different feature engineering strategies. In particular, we tested whether reducing long AEQ to short assertion clauses via text splicing supports model performance. Insights in this study can be used to fine-tune existing pedagogical-alignment classifiers to AEQ materials (or vice-versa) towards building quality personalized educational materials at scale.
\section{Related Works}
NLP models are developed support the evaluation of educational materials. This suite of models includes feature-based ML models, transformers and advanced LLMs. While each generation of models has improved in-domain classification for cognitive depth (Bloom's Taxonomy) and language proficiency (CEFR), their susceptibility to out-of-distribution (OOD) performance degradation remains a central challenge.

\subsection{Feature-based ML models}
Early automated assessment models relied on expert-crafted linguistic features \citep{benedetto2020} such as text statistics (word counts), Flesch-Kincaid readability index \citep{flesch1948new}, syntactic complexity (Part-of-Speech tags, parse tree depth) \citep{mohammed2020}, and lexical diversity metrics. These features serve as inputs to ML model classifiers such as Multi-nomial Logistic Regression, Tree-based models, and Support-Vector Machine to capture statistical relationships of these features to a label (e.g., Bloom level or CEFR).

\citet{mohammed2020} demonstrated the use of new engineered features such as a term weighting method (TFPOS-IDF) and Word2Vec (W2V) embeddings with ML models to classify the Bloom level of open-domain questions. By weighting Part-of-Speech (POS) such as verbs, nouns, and adjectives with weights calculated from Term frequency-inverse document frequency (TF-IDF), \citet{mohammed2020} found high performance of ML models to classify with micro F1-scores up-to 0.87 (TFPOS-IDF). Furthermore, combining W2V embeddings with TFPOS-IDF features further increased performance to 0.90. \citet{maharramov2025} also found that Convolutional Neural Networks (CNN) with FastText embeddings out-performed 26 model-feature combination with macro F1-score of $0.831\pm0.026$. This underscores the role of feature-engineering in improving the performance of assessment models.

\subsection{Transformer models}
Transformers learn sequence-to-sequence relation of words making them powerful models for text classification tasks  \citep{vaswani2017}. Transformer models are found to be robust classifiers for evaluating Bloom level tasks. \citet{waheed2021} reported their proposed transformer model (BloomNet) which out-performed TF-IDF Random Forest model (macro-F1 scores of 0.87 versus 0.71) for IID dataset and (0.67 versus 0.58) for OOD dataset. The performance drop in BloomNet model from IID to OOD dataset (macro-F1 scores of 0.87 to 0.67) highlights that transformers remain vulnerable to dataset shift.

\subsection{Large-Language Models (LLMs)}
Pre-trained LLMs such as GPT-4 and Gemini can potentially replace model training with zero/Few-shot prompting. LLMs (e.g, GPT-4, Gemini). \citet{huber2025} evaluated different LLMs performance on Bloom level classification task. They found that LLama and GPT-4 show high agreement with human annotators for low-order Bloom levels (remember, understand, and apply) but noted a drop in performance with higher order skills (analyze, evaluate, create). \citet{kumar2025} also found that LLMs (OpenAI GPT-4o-mini and Google Gemini-1.5-pro) had an acceptable performance, macro F1-score $\sim 0.72$. Thus, LLMs offer an out-of-the box assessment model. 
\section{Methodology}
We investigate the performance of ML, transformers and LLMs to assess the Bloom Taxonomy level particularly on new AEQ. We prepared two classes of datasets: Identically Independent Distributed (IID) and Out-of-Distribution (OOD) datasets following the notation of \cite{waheed2021}. The IID dataset is comprised of the collection of questions from text and websites compiled by \cite{lau2025}. The OOD datasets were produced using AEQ Generation from works of \cite{scaria2024} and this study (Asyncform, referred to as "AF" here on). Baseline models were trained using the IID datasets only. The models transfer learning performance were evaluated using the OOD datasets.

We explored several configuration of inputs features for the model such as the TFIDF vectors, TFIDF with NLP metrics and CEFR level, and DistilBERT embeddings. We analyzed the extent of the misclassification using accuracy metrics, word cloud, and local explanable features. We then investigated approaches to improve model performance by feature engineering, text splicing, and retraining.

\subsection{Asyncform (AF)} In this study, we generated 863 AEQ covering subjects in English, Mathematics, and Science. AF dataset was labeled with corresponding Bloom Taxonomy level by three K-12 teachers who taught the subjects. Content information (i.e., subject, grade level, learning objectives, and question) was presented to the teachers in the annotation process. Due to limited staff, only a single Bloom label was provided for a given AEQ. 

\subsection{Feature Engineering}\label{sec:feateng} To mitigate OOD performance degradation, we tested four feature-engineering configurations (Configurations A-D) that capture syntactic, semantic, and Bloom signals.

Configuration A (TFPOS-IDF vector) adapts the approach of  \citep{mohammed2020} which uses a TFPOS-IDF feature to represent text using its term frequency with part-of-speech (POS) weighting that highlights the impact of verbs, nouns, and adjectives in Bloom level classification task. As a pre-processing step, words were stemmed and weighted based on their POS-tags, i.e., 5 for verbs, 3 for nouns or adjectives, and 1 for others. The processed text was then converted to a sparse-vector representation using TF-IDF vectorizer trained on IID vocabulary. 

Configuration B (TFPOS-IDF + NLP Metrics + CEFR level) extends the TFPOS-IDF to include text metrics and CEFR level. We hypothesize that these metrics could improve classification performance.

Configuration C (Text + LO) concatenates the learning objectives (LO), when available, to the question. This is applicable to the AF dataset. The joined text (question + LO) is then processed following TFPOS-IDF approach in Configuration A.

Configuration D utilizes adapts the approach of \citep{kumar2025} which utilized pre-trained DistilBERT embeddings as input features. Rather than using the $\mathit{[CLS]}$ token, we used the pooled attention weights (768 x 1 vector) to train the downstream Fully-Connected-Network (FCN) classifier.

\subsection{Models} We used ML, transformer, and LLMs for the Bloom classification task. The ML models include: Multi-nomial Logistic Regression, Random Forest (RF), Support Vector Machine (SVM), and Fully Connected Network (FCN). We implemented standard model architecture: RF (100 estimators), SVM (linear kernel), XGB (100 estimators), and FCN (128x128 hidden layers). We set up a  DistilBERT uncased transformer model \citep{sanh2019} with an FCN (128x128 hidden layers, 0.30 drop-out and RELU activation function). For LLM evaluation, we used OpenAI GPT-4.1 and Google Gemini Flash 3.1 to estimate the Bloom level using zero-shot and few-shot prompting (see Annex). 

\subsection{Experimental setup} We evaluated all models using 5-fold cross-validation ($K=5$) along with the full-dataset, reporting the mean and standard deviation across folds. To handle target class imbalance, we incorporated class weighting into the loss-functions of the models while maintaining default hyperameters and freezing pre-trained BERT weights. Light ML models were trained on a Mac M1 computer running with Python 3.11 while the BERT model was trained using Google Colab (A100 GPU) with Adam optimizer  ($\mathit{learning\_rate}=1\times 10^{-3}$) for 50 epochs and fixed random seed of 42.

\section{Experiments}
\subsection{Textual characteristics of IID and OOD} We measured the textual characteristics of the dataset such as: text statistics, readability metrics, syntactic depth, and CEFR level estimate.

A set of metrics was used to describe the textual characteristics of the questions namely: text length (number of words), Flesch-Kincade Grade level, number of sentences, maximum syntactic depth, and number of root children. These metrics were calculated using Textstat \citep{ward2022} and Spacy \citep{spacy2020} Python libraries.

A BERT model pre-trained with CEFR-SP dataset \citep{arase2022} was used to estimate the CEFR level of the IID and OOD dataset (see Annex). The predicted CEFR level was used as additional feature in exploring different feature configuration.

\subsection{Feature Engineering and Text Splicing} We measure the classification performance via macro F1-score of different models on the IID and OOD datasets. We also examined potential improvement in the F1-score with feature engineering (see Section~\ref{sec:feateng}) and text splicing.

We hypothesize that splicing text into sentences allows the ML models to locally detect the relevant Bloom triggers (Verbs, Nouns). In text splicing, a full text question, $Q_i$, is split based on punctuations resulting to a set of single sentences, $\{q\}_i$.
$$
   Q_i = \{q_1, q_2, \ldots, q_n\}_i 
$$

The set of questions serves as input to pre-trained models to evaluate the Bloom level per sentence $\{B_1, B_2,\ldots, B_n \}_i$. The maximum Bloom level in the set represented the Bloom level for the question, $Q_i$,
$$
    B_i = \max \left[ \{b_1, b_2, \ldots, b_n\}_i \right]
$$

\subsection{Misclassification and Improvements} We focused our analysis on the best performing ML models from Section~\ref{sec:modelperf}. We hypothesized that misclassification is due to the dilution and/or lack of the Bloom signal in the text. We characterize the signal dilution as the drop in F1-score with text length.
\section{Results}
\subsection{Textual Characteristics of IID and OOD.} Table \ref{tab:datasetchar} shows the overall textual characteristics of the IID and OOD datasets. The IID dataset compiled by \cite{lau2025} has a large sample size $N=6,175$ compared to the total OOD dataset of $N=2,696$ entries. 

\begin{table}[ht]
\centering
\caption{Textual Characteristics of IID and OOD Datasets}
\label{tab:datasetchar}
\begin{tabular}{lccc}
\hline
\textbf{Metric} & \textbf{IID Lau} & \textbf{OOD Scaria} & \textbf{OOD AF} \\
\hline
$N$ & 6,175 & 1,833 & 848 \\
length (mean) & 9.3 & 27.5 & 18.2 \\
FKG Level (mean) & 9.6 & 14.1 & 9.3 \\
$>$ 1 sentence (\%) & 0.0 & 43.2 & 17.3 \\
$>$ 2 clauses (\%) & 0.0 & 66.8 & 51.4 \\
$>$ 5 max depth (\%) & 34.9 & 94.0 & 64.5 \\
$>$ 2 root children (\%) & 13.9 & 65.1 & 89.5 \\
CEFR (A1, A2) (\%) & 2.4 & 0.0 & 0.6 \\
CEFR (B1, B2) (\%) & 88.8 & 63 & 74.5 \\
CEFR (C1, C2) (\%) & 8.9 & 37 & 24.9 \\
\hline
\end{tabular}
\end{table}

IID entries are short (mean length $\sim$ 10 words), single sentences, with shallow syntactic depth (only 34.9\% of data have >5 max depth) and suitable for Grade 10 (FKG ~ 9.6). Majority (88.8\%) of the IID have CEFR levels of B1 and B2. In contrast, the OOD dataset exhibits longer text (mean length > 20 words), with more than one sentence (43\% for Scaria, 17\% for the AF), has deeper syntactic depth (94\% of data have $>5$ max depth) and suitable for Grade levels 9 to College. The OOD estimated English proficiency level are higher covering B1-B2 (at least 63\% of Scaria) and C1-C2 (at least 25\% of AF).

\begin{figure}[htbp!]
\centering
\includegraphics[width=0.90\textwidth,keepaspectratio]{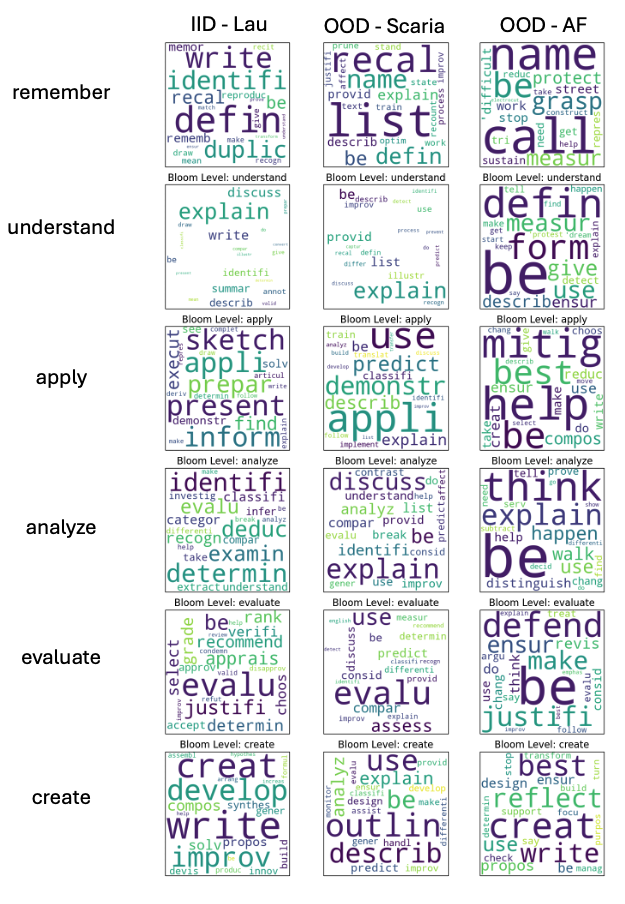}
\caption{The top 20 frequently used verbs across different Bloom levels in each datasets.}
\label{fig:wordcloud}
\end{figure}

Word cloud analysis shows that IID dataset contains rich sets of verbs in different Bloom levels (see Figure \ref{fig:wordcloud}). However, the OOD dataset shows only partial overlap on the set of Bloom trigger verbs. The OOD-Scaria dataset shows the highest overlap of Bloom Verbs (median = 35.1\%) while the AF dataset shows least overlap (median = 10.5\%).

\subsection{Model performance with Feature Engineering and Text Splicing}\label{sec:modelperf} 
There is a low transfer performance between the IID and OOD test datasets (see Table \ref{tab:modelperformance}). A Naïve model - with the majority class as the prediction, has macro F1-scores of 0.09 (IID) and 0.06 (OOD-Scaria) and 0.07 (OOD-AF). With TFPOS-IDF as features, all models show high F1-score (~0.88) with the IID datasets, particularly XGBoost. However, their performance decreased with the OOD dataset to $0.48\pm0.02$ (Scaria, XGBoost) and $0.18\pm0.02$ (AF, XGBoost).

The addition of NLP and CEFR as features showed marginal improvement in the F1-score particularly for RF model from (0.20 to 0.25 in the OOD-AF).Complex models such as DistilBERT and LLMs show higher performance compared to previous ML models. DistilBERT model had F1-scores of 0.89 (IID) and 0.35 to 0.56 (for OOD). LLMs (GPT4.1 and Gemini-Flash 3.1 lite) showed robust classification performance with the OOD datasets, F1-scores from 0.41 to 0.79.

\begin{table}[ht]
\centering
\caption{Model Performance Comparison: F1 Macro Scores across IID and OOD Datasets}
\label{tab:modelperformance}
\begin{tabular}{lcccc}
\hline
\textbf{Features} & \textbf{Models} & \textbf{IID-Lau} & \textbf{Scaria} & \textbf{OOD-AF} \\
\hline
None & Naïve & 0.09 & 0.06 & 0.07 \\
\hline
TFPOS-IDF & LR & 0.83 $\pm$ 0.01 & 0.39 $\pm$ 0.01 & 0.20 $\pm$ 0.02 \\
& RF & 0.89 $\pm$ 0.01 & 0.44 $\pm$ 0.02 & 0.22 $\pm$ 0.02 \\
& SVM & 0.83 $\pm$ 0.01 & 0.40 $\pm$ 0.01 & 0.22 $\pm$ 0.01 \\
& FCN & 0.79 $\pm$ 0.01 & 0.33 $\pm$ 0.05 & 0.19 $\pm$ 0.02 \\
& XGBoost & 0.88 $\pm$ 0.01 & 0.48 $\pm$ 0.02 & 0.20 $\pm$ 0.04 \\
\hline
TFPOS-IDF & LR & 0.83 $\pm$ 0.01 & 0.40 $\pm$ 0.01 & 0.20 $\pm$ 0.01 \\
+ NLP & RF & 0.87 $\pm$ 0.01 & 0.41 $\pm$ 0.02 & 0.26 $\pm$ 0.05 \\
+ CEFR & SVM & 0.84 $\pm$ 0.01 & 0.40 $\pm$ 0.01 & 0.22 $\pm$ 0.02 \\
& FCN & 0.79 $\pm$ 0.01 & 0.32 $\pm$ 0.04 & 0.20 $\pm$ 0.02 \\
& XGBoost & 0.88 $\pm$ 0.01 & 0.47 $\pm$ 0.01 & 0.23 $\pm$ 0.03 \\
\hline
DistilBERT & FCN & \textbf{0.89 $\pm$ 0.01} & \textbf{0.55 $\pm$ 0.02} & \textbf{0.30 $\pm$ 0.03} \\
\hline
None & Gemini31F-ZS & 0.73 & 0.45 & 0.44 \\
& GPT4.1-ZS & 0.67 & 0.77 & 0.51 \\
& Gemini31F-FS & 0.76 & 0.79 & 0.41 \\
& GPT4.1-FS & 0.72 & 0.79 & 0.47 \\
\hline
\end{tabular}
\end{table}

Text splicing improved the performance of ML and BERT in the Scaria dataset (see Table \ref{tab:modelsplice}). In particular, XGBoost and BERT models F1-scores increased to 0.59 and 0.62, respectively.

\begin{table}[ht]
\centering
\caption{Model Performance with Splicing (SP): F1 Macro Scores across IID and OOD Datasets}
\label{tab:modelsplice}
\begin{tabular}{llccc}
\hline
\textbf{Features} & \textbf{Models} & \textbf{IID-Lau} & \textbf{Scaria} & \textbf{OOD-AF} \\
\hline
TFIDF + SP & LR & 0.83 $\pm$ 0.01 & 0.41 $\pm$ 0.01 & 0.22 $\pm$ 0.01 \\
& RF & 0.88 $\pm$ 0.01 & 0.56 $\pm$ 0.02 & 0.22 $\pm$ 0.01 \\
& SVM & 0.84 $\pm$ 0.01 & 0.42 $\pm$ 0.01 & 0.24 $\pm$ 0.01 \\
& FCN & 0.79 $\pm$ 0.01 & 0.37 $\pm$ 0.04 & 0.19 $\pm$ 0.04 \\
& XGBoost & 0.88 $\pm$ 0.01 & \textbf{0.59 $\pm$ 0.02} & 0.22 $\pm$ 0.04 \\
\hline
DistilBERT + SP & FCN & 0.89 $\pm$ 0.01 & \textbf{0.62 $\pm$ 0.03} & \textbf{0.29 $\pm$ 0.02} \\
\hline
\end{tabular}
\end{table}

\newpage
\subsection{Misclassifications}\label{sec:misclass}
The Figure \ref{fig:cmbaseline} below shows the confusion matrix for XGBoost, BERT, and Gemini-3.1-Flash revealing model biases across datasets. For the IID dataset, XGBoost and BERT perform well across Bloom Levels. Gemini showed a high number of misclassifications particularly on low order Bloom levels ("remember" and "understand"). For the OOD dataset, both pre-trained XGBoost and BERT models show a bias in "understand" prediction in Scaria and AF dataset. BERT model correctly predicts higher-order Bloom level ("create") questions in the Scaria dataset. In contrast, Gemini-FS shows robust performance with relatively high accuracy across bloom levels.

\begin{figure}[htbp!]
\centering
\includegraphics[width=1.0\textwidth,keepaspectratio]{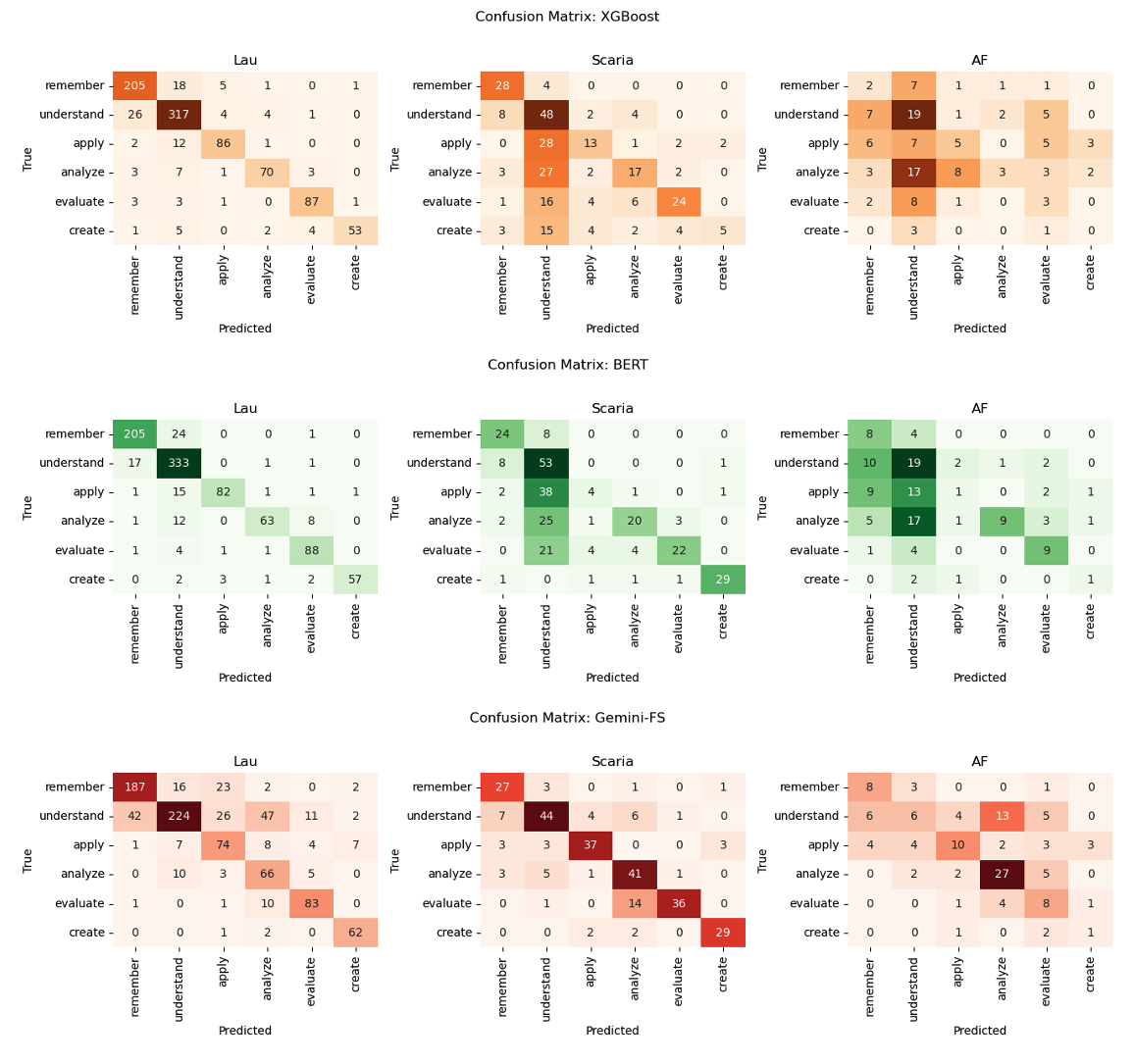}
\caption{The confusion matrix of the XGB, DistilBERT, and Gemini-3.1-Flash with few-shot prompting on the IID and OOD dataset.}
\label{fig:cmbaseline}
\end{figure}

In terms signal dilution, model accuracy tends to drop with text length. However, the extent of dilution varies with IID and OOD datasets. For IID-Lau, the accuracy of BERT model drops to 80\% with text length 20-30 words. For OOD-Scaria, performance of ML and BERT models drop to 50\% with text length 15-20 words. LLMs still exhibit dilution however their accuracy remained high (>70\%) even for longer text (text length $>$ 20 words). For OOD-AF, all models show low performance (accuracy < 50\%) for text length 5 - 30 words with LLMs having the highest score.

\begin{figure}[htbp!]
\centering
\includegraphics[width=1.0\textwidth,keepaspectratio]{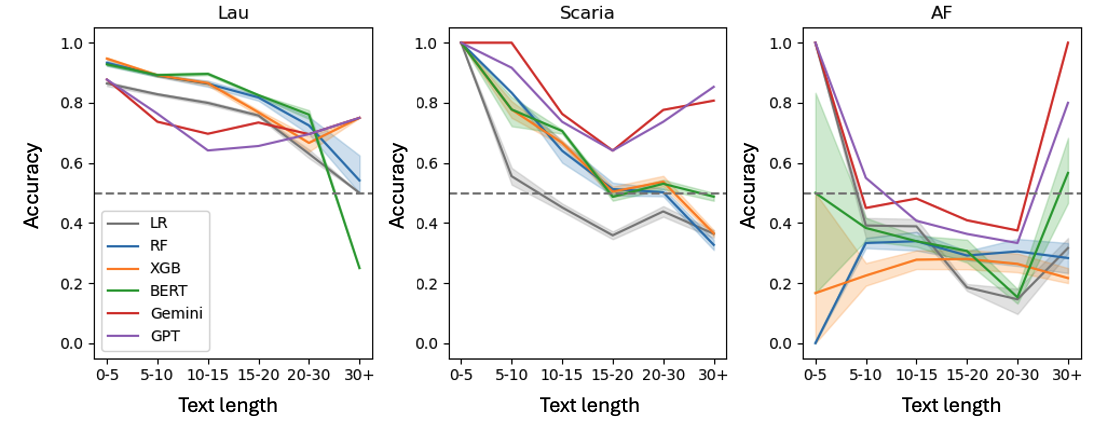}
\caption{The drop in the accuracy of different models with the text length across the IID and OOD dataset indicating signal dilution.}
\label{fig:dilution}
\end{figure}

\newpage

\subsection{Improvements summary}
\begin{figure}[htbp!]
\centering
\includegraphics[width=1.0\textwidth,keepaspectratio]{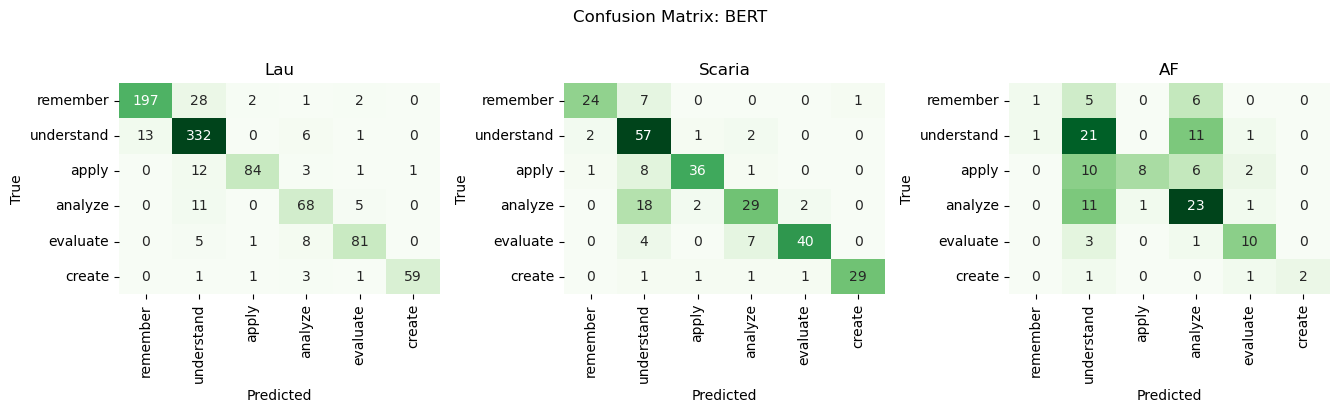}
\caption{The confusion matrices of the retrained BERT models across the IID and OOD datasets.}
\label{fig:CMRet}
\end{figure}

The tables below summarizes the macro F1-scores of ML and BERT to the OOD datasets following the addition of features, learning objective, text splicing, and retraining. The baseline case show that DistilBERT having the best performance in Scaria \ref{tab:improvscaria} and AF dataset \ref{tab:improvaf}. In general, improvements were observed following text splicing, addition of learning objectives and model retraining. With the Scaria dataset, XGBoost model performance increased with text splicing (0.59 versus 0.48 from baseline). With retraining, XGBoost model also out-perform BERT model (0.82 versus 0.77). With the AF dataset, BERT was better performing than XGBoost. BERT performance improved following retraining (0.50 versus 0.30) at par with the best performing LLMs for AF dataset (0.51, GPT4.1 - ZS). Model retraining addressed model biases  on "understand" to "analyze" bloom levels (see Figure \ref{fig:CMRet}) particularly for Scaria dataset. However, prediction bias is still evident with the AF dataset. The inclusion of learning objectives in the text during retraining marginally improved BERT performance (0.54 versus 0.50). 

\begin{table}[htbp!]
    \centering
    \caption{Performance on Scaria Dataset}
    \label{tab:improvscaria}
    \begin{tabular}{l|ccc}
        \hline
        \textbf{Method} & \textbf{LR} & \textbf{XGBoost} & \textbf{BERT} \\
        \hline
        Baseline & 0.39 & 0.48 & 0.55 \\
        + Text Metrics & 0.40 & 0.47 & NA \\
        + Splice & 0.41 & \textbf{0.59} & \textbf{0.62} \\
        Retrain & 0.71 & \textbf{0.82} & \textbf{0.77} \\
        \hline
    \end{tabular}
\end{table}

\begin{table}[htbp!]
    \centering
    \caption{Performance on AF Dataset}
    \label{tab:improvaf}
    \begin{tabular}{l|ccc}
        \hline
        \textbf{Method} & \textbf{LR} & \textbf{XGBoost} & \textbf{BERT} \\
        \hline
        Baseline       & 0.19 & 0.20 & 0.30 \\
        + Text Metrics & 0.19 & 0.23 & NA \\
        + LO           & 0.25 & 0.20 & \textbf{0.40} \\
        + LO + Splice  & 0.24 & 0.33 & \textbf{0.36} \\
        + Retraining   & 0.43 & 0.36 & \textbf{0.50} \\
        + Retraining (with LO) & 0.43 & 0.36 & \textbf{0.54}\\
        \hline
    \end{tabular}
\end{table}

\section{Discussion}
In this study, we evaluated the performance of pre-trained ML-based Classification models applied to novel AI-assisted educational questions OOD datasets. Consistent with the findings of \citep{waheed2021}, the BERT classification models perform well on the datasets they were trained (IID). However, these models struggle to classify novel AI-assisted questions. We suggest two possible explanations: (1) AI-assisted education questions are verbose, which dilutes the Bloom signals; and (2) generated questions lack explicit use of Bloom trigger verbs that are important for classification.

Because AI-assisted questions tend to incorporate context or situational examples, the increased text length dilutes Bloom signal and decreases the performance of TFPOS-IDF classifiers. As shown in Section~\ref{sec:misclass}, model accuracy drops with text length with Bloom Signal dilution specific to different datasets. To enhance the signal-to-noise ratio, text splicing was effective in isolating sentences that contain clear Bloom levels evident in the Scaria Dataset.

AI-assisted Educational questions tend to lack clear bloom signal. Most of pre-trained Bloom classifiers are designed to detect presence of Bloom Trigger Verbs. With the AF dataset, we found that adding the learning objective to the questions substantially improved classification performance. This finding aligns with \citep{li2022}, demonstrating that the learning objectives serve as anchor features for Bloom level classification task . We noted that splicing had a detrimental effect on the classification score. This indicates that the combination of questions and learning objective provide clear Bloom signals.

Model retraining substantially improved model performance. Notably, the performance of the XGBoost model was comparable to BERT models. This indicates that the existing TFPOS-IDF features are sufficient. However, this task requires a large dataset, particularly for complex models. 

The marginal improvements on the re-trained models on the AF-dataset in Table \ref{tab:improvaf} indicate a novel pattern that is not captured by the models. Using LIME interpretability tool, we found muted impact of nouns(NN), verbs(VB), and adjectives(JJ) in the AF Dataset to predicted Bloom Level (LIME weight< 0.50) (see Annex 3). To improve AF classification, we may need to (1) add more training data for the transformer to learn complex relations of NN, VB, and JJ; (2) consider restructuring the questions to include high-impact VB, NN.

From these findings, we provide three recommendations to address loss of model performance in novel OOD datasets. First, if no access to annotated datasets, use BERT and LLMS which provide robust bloom classification performance. Second, if with access to sample OOD datasets with labels, explore feature engineering approaches (e.g., text splicing and addition of learning objectives) to transform OOD dataset to be similar to IID data. Third, if one has access to a large OOD labeled dataset (N > 1,000 samples), retrain the model.
\section{Summary and Future Work}
In conclusion, we demonstrate that AI-generated questions - due to its verbosity introduce dataset shift degrading performance of pre-trained Bloom Taxonomy Classifiers. Our findings align with previous observations that models suffer a performance drop of approximately 20\% when moving to out-of-distribution educational content. We show that techniques like text splicing and the integration of learning objectives provide additional information for "Bloom signals" in classification. While transformer-based models like BERT and LLMs consistently offer more robust baseline performance than traditional frequency-based methods, strategic model retraining remains the most effective path toward achieving high pedagogical alignment. By establishing a validated framework for the assessment of synthetic content, this study enables the scalable production of high-quality learning materials that strictly adhere to established standards such as the Revised Bloom Taxonomy. These results also highlight the need for collaboration between AI developers and educators, ensuring that the surge in automated content generation is met with equally rigorous automated validation.

\newpage
\bibliographystyle{apacite}
\bibliography{references}  






\newpage
\section*{Annex 1. Bloom Taxonomy Zero-Shot Prompt}

\begin{Verbatim}
Context: You are an expert in learning pedagogical classification based on Bloom's Taxonomy. 

Task: Your task is to analyze the given text or narrative and determine its cognitive level
[remember, understand, apply, analyze, evaluate, create] based on the verbs and concepts presented.

Provide only the cognitive level as output, without explanation or justification.

CRITICAL INSTRUCTIONS:
- Output ONLY a valid JSON object with keys "remember", "understand", "apply", "analyze", "evaluate", and "create"
- value must be probability scores (float) for each Bloom level, summing to 1.0
- Do NOT provide explanations or additional text


TEXT: \{document\_text\}

JSON Response:
\end{Verbatim}

\section*{Annex 2. Bloom Taxonomy Few-Shot Prompt}
\begin{Verbatim}
Context: You are an expert in learning pedagogical classification based on Bloom's Taxonomy. 

Task: Your task is to analyze the given text or narrative and determine its cognitive level
[remember, understand, apply, analyze, evaluate, create] based on the verbs and concepts presented.

Provide only the cognitive level as output, without explanation or justification.

Below are some examples:

Example 1:
TEXT: "defend the following claim the cornell method works so well that it could turn even a poor lecture into a valuable learning experience."
OUTPUT: remember: 0.05, understand: 0.10, apply: 0.05, analyze: 0.10, evaluate: 0.65, create: 0.05

Example 2:
TEXT: "why fft is needed"
OUTPUT: remember: 0.15, understand: 0.75, apply: 0.05, analyze: 0.03, evaluate: 0.01, create: 0.01

Example 3:
TEXT: "differentiate between different perspectives on an issue"
OUTPUT: remember: 0.05, understand: 0.1, apply: 0.05, analyze: 0.75, evaluate: 0.05, create: 0.0


CRITICAL INSTRUCTIONS:
- Output ONLY a valid JSON object with keys "remember", "understand", "apply", "analyze", "evaluate", and "create"
- value must be probability scores (float) for each Bloom level, summing to 1.0
- Do NOT provide explanations or additional text


TEXT: \{document\_text\}

JSON Response:
\end{Verbatim}

\newpage
\section*{Annex 3. LIME Interpretability}

\begin{figure}[htbp!]
\centering
\includegraphics[width=0.6\textwidth,keepaspectratio]{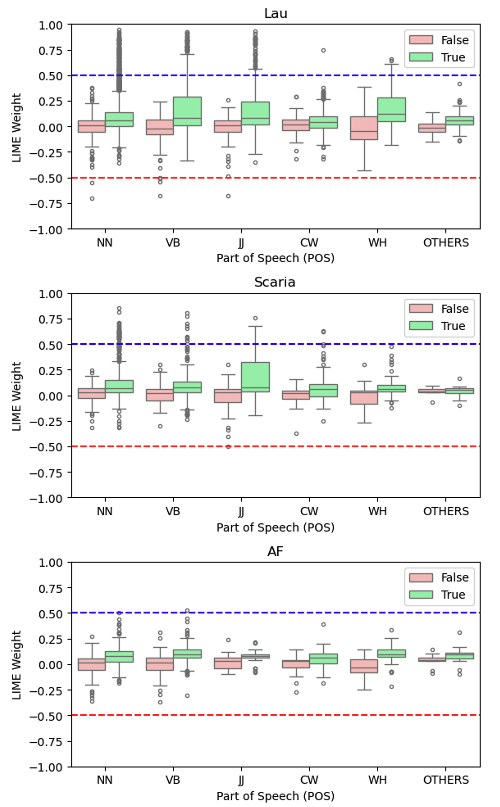}
\caption{The box-plot distribution of the feature importance with the retrained BERT model using LIME. The model relies on high impact nouns, verbs, and adjectives (NN, VB, JJ) to form a correct prediction. The POS distribution also varies across datasets suggesting that each dataset have different text structure which impacts model performance.}
\label{fig:LIME}
\end{figure}

With the retrained BERT model, we analyzed the model feature importance via LIME across the three datasets. Features were grouped by their POS tags and accuracy (see Figure \ref{fig:LIME}).  LIME evaluation show that model relies on presence of high-impact of nouns, verbs, and adjectives (NN, VB, and JJ) for correct prediction across datasets. Across the Lau and Scaria dataset, correct predictions relied on high impact tokens (NN, VB, and JJ with LIME weights > 0.50). In addition, the Scaria dataset included reliance on CW(Connecting and Structural words, e.g., “in”, “the”, “on”, “a”). In contrast, the AF dataset show muted impact of NN, VB, and JJ. Using Chi-square test confirmed high association between dataset domain and high-impact verbs ($\chi^2 = 31.44$, $p < 0.001$) and Nouns ($\chi^2 = 8.54$, $p < 0.014$) but not on adjectives ($p = 0.085$). To improve AF classification, we may need to (1) add more training data (consistent with paper) for the transformer to learn complex relations of NN, VB, and JJ; (2) consider restructuring the questions to include high-impact VB, NN.

\end{document}